%% file: main.tex
\documentclass[10pt,twocolumn,letterpaper]{article}

\usepackage{iccv}              
\definecolor{iccvblue}{rgb}{0.21,0.49,0.74}
\usepackage[pagebackref,breaklinks,colorlinks,allcolors=iccvblue]{hyperref}
\usepackage{pifont}
\usepackage{multirow}
\usepackage{makecell}
\usepackage{diagbox}
\usepackage{utfsym}
\def\paperID{6095} 
\def\confName{ICCV}
\def\confYear{2025}
\usepackage{booktabs}

\title{TR-PTS: Task-Relevant Parameter and Token Selection for Efficient Tuning}

\author{
Siqi Luo$^1$\thanks{Equal Contribution.}~,
~Haoran Yang$^1$\footnotemark[1]~, 
~Yi Xin$^2$\footnotemark[1]~, 
~Mingyang Yi$^3$, 
~Guangyang Wu$^1$,\\
~Guangtao Zhai$^1$, 
~Xiaohong Liu$^{1,4}$\thanks{Corresponding Author.} \\[4pt]
$^1$Shanghai Jiao Tong University \quad
$^2$Shanghai Innovation Institute \quad
$^3$Renmin University of China \\[4pt]
$^4$Suzhou Key Laboratory of Artificial Intelligence \\
{\tt\small \{siqiluo647, yhr-73, wu.guang.young, zhaiguangtao, xiaohongliu\}@sjtu.edu.cn} \\
{\tt\small xinyi@smail.nju.edu.cn} \quad
{\tt\small yimingyang@ruc.edu.cn}
}

\begin{document}
\maketitle
\input{sec/0_abstract}    
\input{sec/1_intro}

\input{sec/2_related}

\input{sec/3_method}
\input{sec/4_experiment}

\input{sec/5_conclusion}

{
    \small
    \bibliographystyle{ieeenat_fullname}
    \bibliography{main}
}
\end{document}

%% file: sec/0_abstract.tex
\begin{abstract}
Large pre-trained models achieve remarkable performance in vision tasks but are impractical for fine-tuning due to high computational and storage costs. Parameter-Efficient Fine-Tuning (PEFT) methods mitigate this issue by updating only a subset of parameters; however, most existing approaches are task-agnostic, failing to fully exploit task-specific adaptations, which leads to suboptimal efficiency and performance. To address this limitation, we propose \textbf{T}ask-\textbf{R}elevant \textbf{P}arameter and \textbf{T}oken \textbf{S}election (TR-PTS), a task-driven framework that enhances both computational efficiency and accuracy. 
Specifically, we introduce \textbf{Task-Relevant Parameter Selection}, which utilizes the Fisher Information Matrix (FIM) to identify and fine-tune only the most informative parameters in a layer-wise manner, while keeping the remaining parameters frozen.
Simultaneously, \textbf{Task-Relevant Token Selection} dynamically preserves the most informative tokens and merges redundant ones, reducing computational overhead. 
By jointly optimizing parameters and tokens, TR-PTS enables the model to concentrate on task-discriminative information. We evaluate TR-PTS on benchmark, including FGVC and VTAB-1k, where it achieves state-of-the-art performance, surpassing full fine-tuning by 3.40\% and 10.35\%, respectively. The code are available at \url{https://github.com/synbol/TR-PTS}.
\end{abstract}

%% file: sec/1_intro.tex
\section{Introduction}
\label{sec:intro}
Pre-trained on large-scale datasets and subsequently fine-tuned for downstream tasks, this paradigm has become the standard framework across various domains, including Natural Language Processing (NLP)~\cite{devlin2018bert, brown2020language}, Computer Vision (CV)~\cite{xin2024v,luo2024enhancing,xin2025lumina,du2025multi,qin2025lumina}, and others~\cite{xin2023mmap,liu2025m2ist}.
Traditionally, fine-tuning a pre-trained model to a specific task involves adjusting all parameters, a method known as full fine-tuning. However, with state-of-the-art models now comprising billions or even trillions of parameters~\citep{openai2023gpt4, touvron2023llama}, this conventional approach has become increasingly impractical due to its exorbitant computational and storage requirements.

\begin{figure}[t]
\centering
\includegraphics[width=0.42\textwidth]{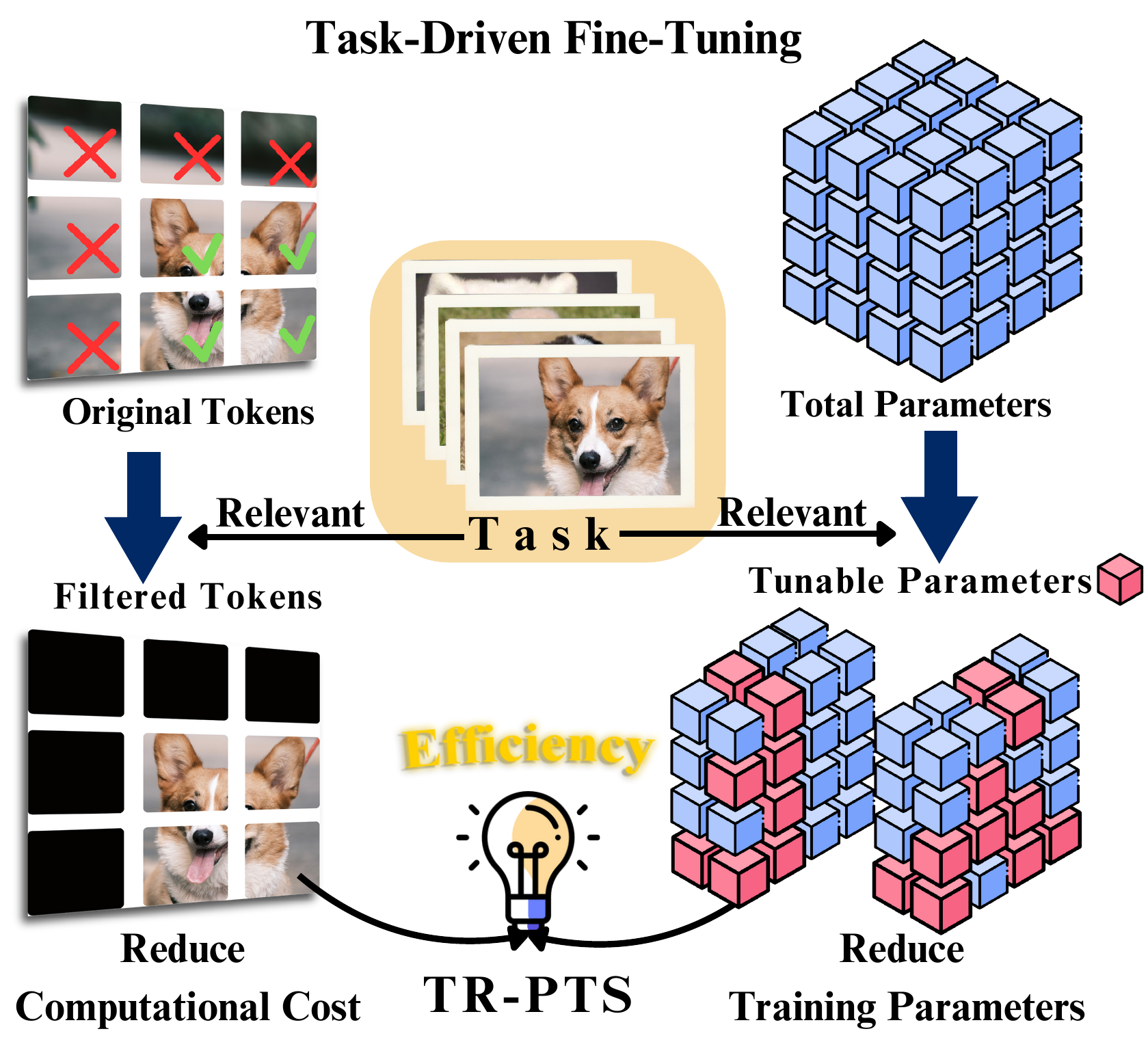}
\caption{\textbf{Overview of Our TR-PTS}. 
TR-PTS selectively retains task-relevant tokens and parameters for efficient fine-tuning.  
}
\vspace{-0.6cm}
\label{fig:intro}
\end{figure}

\begin{figure}[t]
\centering
\vspace{-0.15cm}
\includegraphics[width=0.48\textwidth]{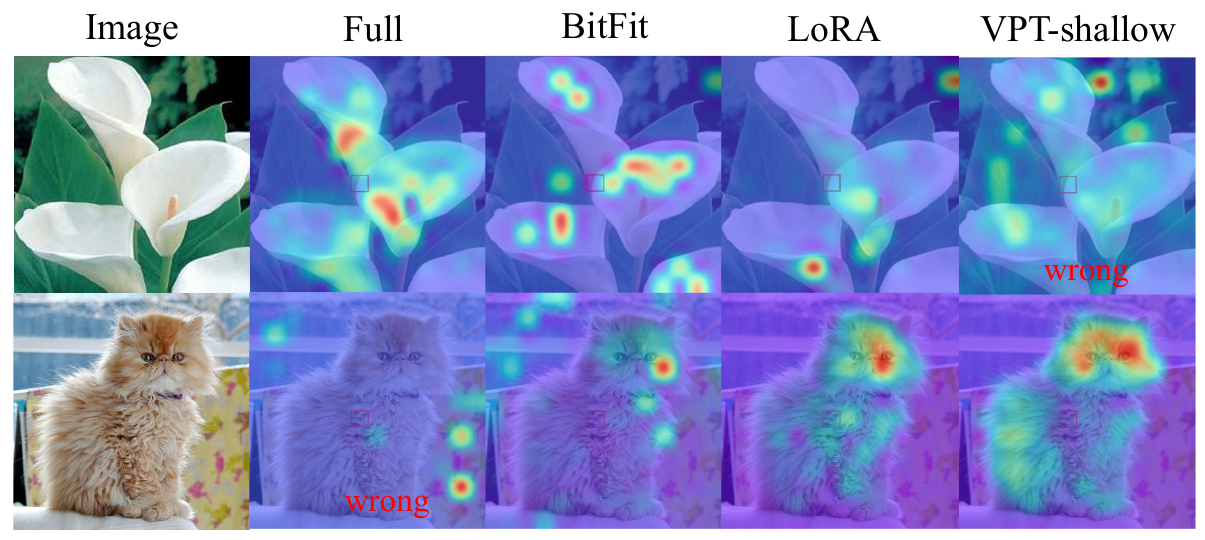}
\vspace{-0.65cm}
\caption{\textbf{Visualizing the Impact of Each Token on the Final Prediction.} The label "wrong" indicates an incorrect classification. These results demonstrate that the final prediction relies primarily on a subset of the most task-relevant tokens.}
\vspace{-0.6cm}
\label{fig:attention}
\end{figure}

Mitigating the inefficiencies of full fine-tuning has led researchers in the CV domain to explore Parameter-Efficient Fine-Tuning (PEFT) techniques~\citep{zaken2021bitfit, zhang2024gradient,xin2024vmt,liu2024sparse}, which update only a subset of model parameters or introduce lightweight adaptation modules~\citep{houlsby2019parameter, jia2022visual, hu2022lora}. However, as summarized in Table~\ref{table: Comparison between different fine-tuning methods}, existing methods still face several challenges:
\textbf{(1) Inference Overhead:} Some methods, such as VPT~\citep{jia2022visual}, introduce additional learnable modules that increase computational cost during inference. Others, such as LoRA~\citep{hu2022lora}, use reparameterization techniques that can be merged into the backbone at inference time and therefore avoid extra cost.
\textbf{(2) Limited Task-Aware Adaptation:} While some methods have begun to incorporate task-specific adaptation, such as GPS~\citep{zhang2024gradient}, many still use uniform tuning strategies that overlook the varying importance of model components across tasks.
\textbf{(3) Disjoint Optimization:} Most approaches optimize parameter selection and token processing separately, although the informativeness of tokens depends heavily on the task, particularly in ViT-based models. As shown in Figure~\ref{fig:attention}, token importance varies across tasks and should be considered during optimization. This motivates the development of a unified solution that considers both task-relevant parameters and informative tokens.


To address these challenges, we propose \textbf{T}ask-\textbf{R}elevant \textbf{P}arameter and \textbf{T}oken \textbf{S}election (TR-PTS), a novel framework that unifies task-driven parameter selection and token refinement. As illustrated in Figure~\ref{fig:intro}, when a task arrives, the model’s adaptation is manifested both in parameter sensitivity and token informativeness. Guided by this principle, our method dynamically identifies the most relevant parameters and tokens for the task at hand, enabling the model to concentrate computation on the most discriminative information while avoiding redundant updates.

\begin{itemize}
    \item \textbf{Task-Relevant Parameter Selection:}
    Rather than relying on gradient magnitudes~\cite{zhang2024gradient}, which can be affected by optimization noise and may not accurately reflect task importance, we employ the Fisher Information Matrix (FIM)~\cite{kirkpatrick2017overcoming} to quantify parameter sensitivity. We then select the most critical parameters and use a layer-wise allocation strategy to distribute trainable connections according to each layer’s task relevance, reducing the number of trainable parameters while preserving adaptability.

    \item \textbf{Task-Relevant Token Selection:}
    We leverage attention scores from the $[CLS]$ token to retain the most informative tokens while merging the rest through weighted averaging. This allows the model to concentrate attention on discriminative content and reduces computational overhead without sacrificing accuracy.
\end{itemize}

Empirical observations during system design revealed a notable correlation between parameter sparsity and token redundancy: layers with fewer task-relevant parameters tend to encode less informative tokens. Leveraging this insight, we formulate a coordinated selection strategy that applies token reduction preferentially to parameter-sparse layers. This unified, task-aware mechanism enhances computational efficiency while preserving representational integrity, thereby achieving a more favorable trade-off between accuracy and resource consumption.

We evaluate TR-PTS across a diverse set of 24 image recognition tasks. Our approach demonstrates state-of-the-art performance compared to other PEFT methods while achieving an optimal balance between performance, trainable parameters, and computational cost, as illustrated in Figure~\ref{fig:intro}. Compared to full fine-tuning, TR-PTS achieves 3.40\% (FGVC) and 10.35\% (VTAB) improvement of the accuracy while tuning only 0.60\% and 0.34\% parameters of the pre-trained model. Additionally, 
the time required for both fine-tuning and the inference process is significantly reduced, as detailed in the referenced Section~\ref{sec_mainexp}.

\begin{table}
    \centering
        \setlength{\tabcolsep}{2.5pt}
        \resizebox{1\linewidth}{!}{
            \begin{tabular}{@{}ccccc@{}}
            \toprule
            Method  & Task Params.  & Task Tokens  & Memory & Train/Inference\\
             & Adaptive  & Adaptive & Efficient  & Efficient\\
             \midrule
            Full~\cite{jia2022visual} & {\ding{55}} &  {\ding{55}} & {\ding{55}} & {\ding{55}}\\
            
            Linear~\cite{jia2022visual} & {\ding{55}} &  \multicolumn{1}{c}{\ding{55}} & \multicolumn{1}{c}{\ding{51}}  & \multicolumn{1}{c}{\ding{51}}                  \\ \midrule
            BitFit~\cite{zaken2021bitfit} & \multicolumn{1}{c}{\ding{51}} &  \multicolumn{1}{c}{\ding{51}}  & {\ding{55}}  & {\ding{55}}                 \\ 
            Adapter~\cite{houlsby2019parameter}  & \multicolumn{1}{c}{\ding{55}}& \multicolumn{1}{c}{\ding{55}}  & {\ding{55}}   & {\ding{55}}                 \\
            
            VPT~\cite{jia2022visual}         &\multicolumn{1}{c}{\ding{55}} & \multicolumn{1}{c}{\ding{55}}   & {\ding{55}}     & {\ding{55}}      \\
            LoRA~\cite{hu2021lora}& \multicolumn{1}{c}{\ding{55}}& {\ding{55}}  & {\ding{55}}   & {\ding{55}}                 \\
            SSF~\cite{lian2022scaling} & \multicolumn{1}{c}{\ding{55}}  & {\ding{55}}    & {\ding{55}}     & {\ding{55}}     \\

             GPS~\cite{zhang2024gradient}  & \multicolumn{1}{c}{\ding{51}}  & {\ding{55}}    & {\ding{55}}     & {\ding{55}} 

            \\ \midrule
             TR-PTS (Ours)   &  \multicolumn{1}{c}{\ding{51}}  & \multicolumn{1}{c}{\ding{51}}  & {\ding{51}}       & {\ding{51}}           \\ \bottomrule
            \end{tabular}
        }
         \caption{
         \textbf{Comparison of Different Fine-Tuning Methods.} Our method adaptively selects task-relevant parameters and tokens, allowing it to be applied across various model architectures without adding extra parameters during either training or inference.
         }
         \vspace{-0.5cm}
        \label{table: Comparison between different fine-tuning methods}
\end{table}

%% file: sec/2_related.tex
\section{Related Work}
\label{sec:related}
\subsection{Visual Efficient Fine-Tuning} In the survey on visual efficient fine-tuning~\cite{xin2024parameter}, existing methods are generally categorized into five primary approaches: 1) \textit{Adapter Tuning} methods~\cite{houlsby2019parameter,chen2022adaptformer} introduce small-scale neural modules (adapters) into Transformer layers, tuning only these adapters for model adaptation; 2) \textit{Prompt Tuning} methods~\cite{jia2022visual} enhance the original input by adding additional visual prompts; 3) \textit{Specification Tuning}~\cite{zaken2021bitfit,zhang2024gradient} directly modifies a specific subset of parameters in the Transformer to improve efficiency; 4) \textit{Reparameterization Tuning} methods~\cite{lian2022scaling,hu2021lora} introduce new learnable parameters during training, which are later integrated into the original Transformer layers through reparameterization during inference; 5) \textit{Unified-based Tuning} methods~\cite{zhang2022neural} provide a unified framework that integrates various fine-tuning techniques into a single, cohesive architecture. 

The method we propose differs fundamentally from these existing approaches, as shown in Table~\ref{table: Comparison between different fine-tuning methods}. While most of the aforementioned methods focus primarily on the structural design of the model, our approach emphasizes the identification and tuning of important parameters and tokens that are specifically relevant to the downstream tasks.

\subsection{Token Reduction for Efficient ViT}
Since the introduction of ViTs, researchers have explored ways to enhance their efficiency by reducing redundant tokens. Existing methods primarily fall into two categories: token pruning~\cite{wang2024zero,fayyaz2022adaptive} and token merging~\cite{bolya2023tokenmergingvitfaster,rao2021dynamicvit}.
Pruning-based approaches remove less important tokens to lower computational complexity and merging-based methods preserve more information by fusing similar tokens, reducing sequence length. 
The method we propose combines the advantages of pruning and merging. We dynamically select task-relevant important tokens for retention. Instead of discarding unimportant tokens, we perform weighted merging based on task relevance.

%% file: sec/3_method.tex
\section{Method}
\label{sec:method}
Our method is built upon Vision Transformer (ViT)~\cite{dosovitskiy2020vit} architecture. In the following, we first review the ViT architecture in Section~\ref{sec:pre}. Subsequently, we present the proposed \textbf{T}ask-\textbf{R}elevant \textbf{P}arameter and \textbf{T}oken \textbf{S}election (TR-PTS) in Section~\ref{sec:params} and Section~\ref{sec:token}. Finally, we present the integration of these components into a unified Task-Driven Fine-Tuning strategy in Section~\ref{sec:finetuning}.
\subsection{Preliminary}
\label{sec:pre}
The standard ViT consists of a patch embedding layer and a stack of Transformer layers. Given an image $I\in \mathbb{R}^{H\times W\times C}$, the patch embedding layer first splits and flatten the image $I$ into sequential patches $I_p\in \mathbb{R}^{N\times(P^2C)}$, where $(H, W)$ represents the height and width of the input image, $(P, P)$ is the resolution of each image patch, $C$ denotes the number of channels, and $N=HW/P^2$ is the number of image tokens. Then, the patches $I_p$ are mapped to $X_0 = [x_1, x_2, ..., x_N]\in \mathbb{R}^{N\times d}$ using a trainable linear projection. The inputs to the Transformer layers consist of $X_0$ along with a prepended $[CLS]$ token. 
\begin{figure*}[t]
\centering
\vspace{-0.35cm}
\includegraphics[width=0.85\textwidth]{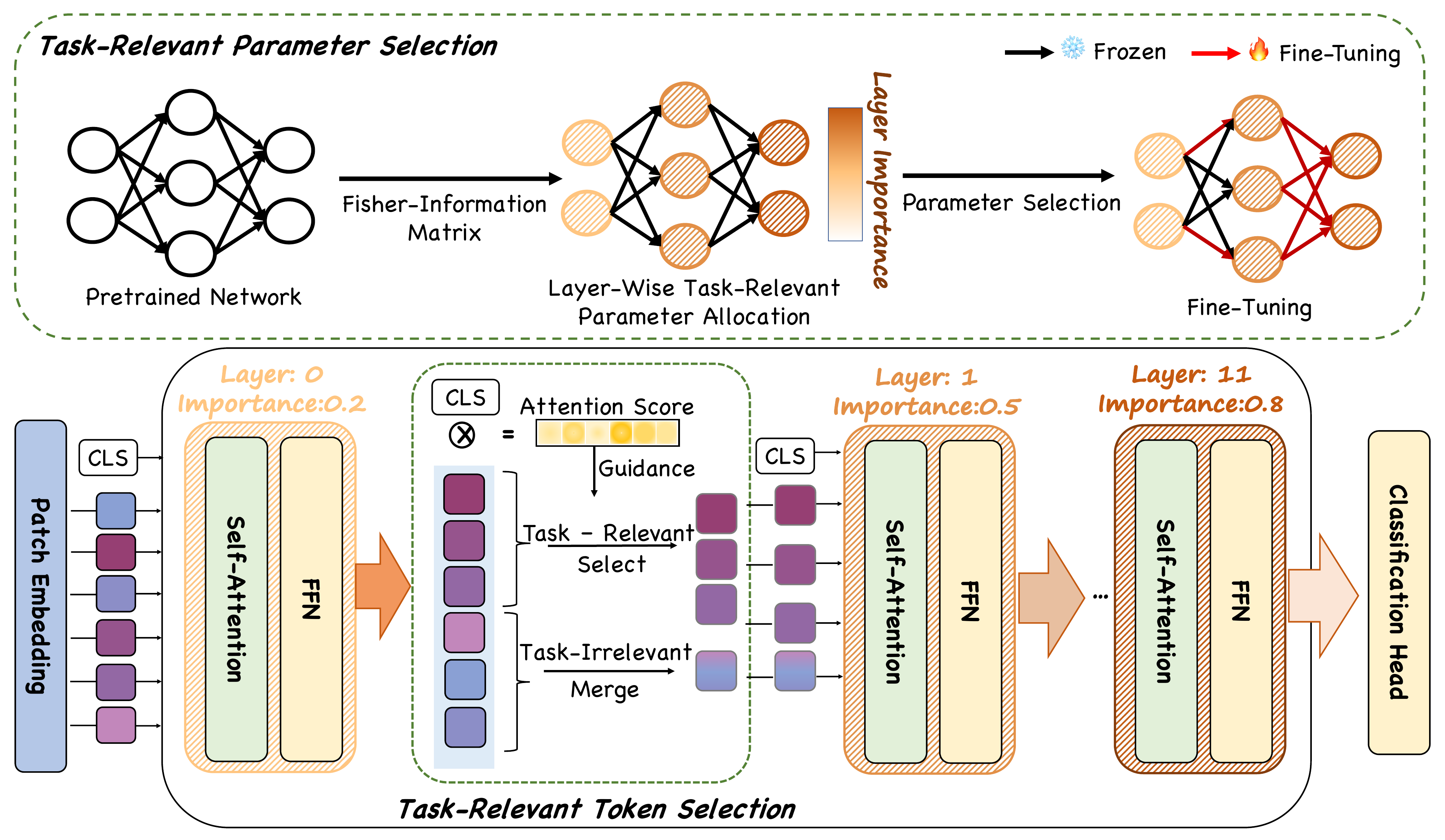}
\vspace{-0.25cm}
\caption{\textbf{Architecture of Our Proposed TR-PTS Framework.}
\textbf{Top:} For Task-Relevant Parameter Selection, we use the Fisher Information Matrix to determine each layer's importance and dynamically set the number of trainable connections per neuron (\(N_l\)); critical layers receive more parameters, while every layer updates at least one connection.
\textbf{Bottom:} For Task-Relevant Token Selection, attention scores from the $[CLS]$ token identify key tokens, and less informative tokens are merged via weighted averaging. Trainable parameters are allocated per layer in proportion to their Fisher scores.}
\label{fig:pipeline}
\vspace{-0.25cm}
\end{figure*}

\subsection{Task-Relevant Parameter Selection}
\label{sec:params}
Existing gradient-based parameter selection methods~\cite{zhang2024gradient} often fail to capture task relevance effectively, leading to uniform parameter selection across layers. To address this, we introduce a layer-wise selection strategy leveraging the Fisher Information Matrix (FIM)~\cite{kirkpatrick2017overcoming}, which quantifies each parameter’s contribution to task-specific adaptations.


\subsubsection{Fisher Information Matrix (FIM)}
FIM evaluates the sensitivity of model outputs to perturbations in parameters. Higher FIM values indicate stronger task relevance, suggesting priority for fine-tuning. Formally, the FIM for model parameters \(\theta\) is:
\vspace{-0.15cm}
\begin{equation}
\mathcal{F}(\theta) = \mathbb{E} \left[ \left( \frac{\partial \log p(y|x; \theta)}{\partial \theta}\right) \left( \frac{\partial \log p(y|x; \theta)}{\partial \theta}\right)^\top \right],
\vspace{-0.2cm}
\end{equation}
where \( p(y|x; \theta) \) represents the conditional probability of output \( y \) given input \( x \). This matrix characterizes the covariance of the gradient of the log-likelihood with respect to the model parameters, thereby identifying the parameters that exert the most substantial influence on task performance.

In practice, we approximate the FIM using the gradients of the cross-entropy loss function \(\mathcal{L}_{\text{CE}}\) in image classification tasks,  as its gradient aligns with the first-order derivatives of the log-likelihood function. Specifically, we use the squared gradients to approximate the diagonal of the FIM:
\vspace{-0.1cm}
\begin{equation}
\mathcal{F}(\theta) \approx \mathbb{E}_{(x, y) \sim D} \left[ \left( \frac{\partial \mathcal{L}_{\text{CE}}}{\partial \theta} \right)^2 \right].
\end{equation}

This approximation avoids the complexity of explicitly computing second-order derivatives while providing an efficient estimate of parameter importance in relation to the target task.

\subsubsection{Layer-Wise Parameter Allocation}

\paragraph{Task-Relevant Layer Importance.}
Instead of selecting parameters uniformly across layers, we introduce a layer-aware allocation strategy to ensure balanced adaptation. We compute task-driven layer-wise importance scores using the Fisher Information Matrix (FIM), allowing us to allocate trainable parameters based on their task relevance.

Once the FIM is computed for all parameters, we first select the top \( M\% \) of parameters with the highest FIM values. Let \( I_{\text{top-M}} \) denote this subset of selected parameters. The contribution of each layer \( l \) is computed as:
\vspace{-0.15cm}
\begin{equation}
    w_l = \frac{|I_{\text{top-M}} \cap L_l|}{|I_{\text{top-M}}|},
    \vspace{-0.15cm}
\end{equation}
where \( L_l \) represents the parameters in layer \( l \). The computed \( w_l \) values determine the relative importance of each layer, ensuring that task-relevant layers receive more trainable parameters.

\vspace{-0.15cm}
\paragraph{Adaptive Layer-Wise Parameter Selection.}
Inspired by the Gradient-based Parameter Selection (GPS) method~\cite{zhang2024gradient}, we introduce a selection strategy that ensures each neuron in layer \( L \) fine-tunes exactly \( C_l \) connections. To dynamically allocate trainable connections per layer, we normalize \( w_l \) relative to the least important layer:
\vspace{-0.15cm}
\begin{equation}
C_l = \max \left( 1, \frac{w_l}{\min(w)} \cdot C_{\text{min}} \right),
\end{equation}
where \( C_{\text{min}} \) represents the minimum number of selected connections per neuron in the least important layer, and \( \min(w) \) denotes the smallest observed importance ratio. This formulation guarantees that every layer retains at least one active connection while allocating a greater number of parameters to task-critical layers.

Within each layer \( l \), we further refine the selection by identifying a subset of task-relevant connections per neuron based on their Fisher Information Matrix (FIM) scores:
\vspace{-0.1cm}
\begin{equation}
    S_l = \left\{\theta_i \mid i \in \operatorname{arg\,top-} C_l \left( \mathcal{F}(\theta_i) \right), \theta_i \in \mathcal{N}_l \right\},
    \vspace{-0.1cm}
\end{equation}
where \( \mathcal{N}_l \) denotes the set of all input connections to neurons in layer \( l \). This selection strategy ensures that each neuron retains a subset of high-FIM parameters, preventing any part of the network from becoming entirely inactive.







\paragraph{Final Task-Relevant Parameter Set.}
Given the neuron-level selection \( S_n \) and the dynamically determined per-layer \( C_l \), the final task-relevant parameter set is:
\vspace{-0.1cm}
\begin{equation}
    \Theta_{\mathcal{T}} = \bigcup_{l} S_l, \quad \text{where } |S_l| = C_l \cdot |\mathcal{N}_l|,
    \vspace{-0.1cm}
\end{equation}
Here, $|\cdot|$ denotes the cardinality of a set, i.e., the number of elements it contains. By focusing on task-relevant parameters at both the layer and neuron levels, our method minimizes redundant computations while ensuring fine-tuning efficiency.

\subsection{Task-Relevant Token Selection}
\label{sec:token}


In Vision Transformers (ViTs), images are split into patch tokens processed by self-attention, where token importance is inherently influenced by parameter distribution. Our Task-Relevant Parameter and Token Selection (TR-PTS) framework optimizes attention by selecting the most informative parameters, which in turn adaptively guides token selection.

To further enhance computational efficiency, we perform Token Selection and Token Merge progressively. Specifically, token selection is strategically applied to layers with sparse task-relevant parameters to ensure a balanced trade-off between parameter fine-tuning and token efficiency. At each refining layer, we retain only the most task-relevant tokens while merging the less informative ones into a single token. This approach minimizes redundant computation and allows the model to focus on the most crucial tokens, potentially improving overall performance. 

\subsubsection{Token Selection}

In Vision Transformers (ViTs), the $[CLS]$ token acts as a global feature aggregator, attending to all image tokens across layers to accumulate task-discriminative information for classification.

In self-attention, tokens contribute differently to the final output, with attention scores reflecting their importance. Tokens with higher scores influence classification more, while lower-weighted ones carry less discriminative information.

Formally, in a Transformer layer, the self-attention mechanism computes attention scores using the scaled dot-product formula:
\vspace{-0.1cm}
\begin{equation}
    A = \text{softmax} \left(\frac{QK^T}{\sqrt{d}} \right),
\end{equation}
where \( Q \) and \( K \) are the query and key matrices of all tokens, and \( d \) is the key dimension. The attention score \( a_i \) of an image token \( x_i \) relative to the $[CLS]$ token is given by:
\vspace{-0.1cm}
\begin{equation}
    a_{i} = \frac{\exp(q_{\text{CLS}} \cdot k_i)}{\sum_{j=1}^{N} \exp(q_{\text{CLS}} \cdot k_j)},
\end{equation}
where \( q_{\text{CLS}} \) is the query vector of the $[CLS]$ token, and \( k_i \) is the key vector of token \( x_i \). This score quantifies how much information token \( x_i \) contributes to the final $[CLS]$ representation, making it a reliable indicator of task relevance.

To retain the most informative tokens, we introduce a select rate \( \rho \in (0,1] \), which determines the fraction of tokens to be preserved. Specifically, we retain the top \( \lfloor \rho N \rfloor \) tokens based on their attention scores:
\vspace{-0.1cm}
\begin{equation}
    X_{\text{selected}} = \left\{ x_i \mid i \in \operatorname{arg\,top-} \lfloor \rho N \rfloor (a) \right\}.
\end{equation}

By adaptively selecting tokens based on the attention distribution rather than a fixed number, our approach ensures that token selection remains flexible across different layers and tasks. This not only reduces redundant computation but also maintains high task performance.
\subsubsection{Token Merge}
While Token Selection removes redundant tokens, Token Merge ensures that information from discarded tokens is preserved rather than entirely ignored. Instead of discarding low-attention tokens, we merge them into a single aggregated token using a weighted averaging process.

Let \( \mathcal{I} \) be the set of less informative tokens, i.e., tokens that were not selected based on attention scores. We compute a fused token \( x_{\text{merged}} \) as:
\begin{equation}
    x_{\text{merged}} = \frac{\sum_{i \in \mathcal{I}} a_i x_i}{\sum_{i \in \mathcal{I}} a_i}.
\end{equation}

At each refining layer, token selection and merging are performed to progressively refine the token sequence.
This merged token is then appended to the selected tokens to form the refined token sequence:

\begin{equation}
    X_{\text{refined}} = \{x_{\text{CLS}},  X_{\text{selected}}, x_{\text{merged}}\}.
\end{equation}

By merging the discarded tokens into a single representation, we retain global information while significantly reducing the number of tokens processed in subsequent layers. 
This progressive refinement reduces the token sequence length as the model deepens, lowering both computation and memory usage.

\subsection{Task-Driven Fine-Tuning}
\label{sec:finetuning}
After identifying task-relevant parameters (see Section~\ref{sec:params}) and task-relevant tokens (see Section~\ref{sec:token}), we integrate both components into a cooperative fine-tuning framework, where token selection and parameter optimization interactively reinforce each other to enhance task adaptation.
Let \(\Theta\) denote the full set of pre-trained model parameters, and \(\Theta_{\mathcal{T}}\) represent the subset identified as critical for the target task.
To ensure that only the most relevant parameters are updated, we introduce a binary mask \(M\) over \(\Theta\):
\begin{equation}
M_i = \begin{cases} 
1, & \text{if } \theta_i \in \Theta_{\mathcal{T}}, \\
0, & \text{otherwise.}
\end{cases}
\end{equation}
The model is trained by minimizing the loss function \(\mathcal{L}\), computed based on the refined token representation:
\begin{equation}
\min_{\Theta} \mathcal{L}\bigl(f(X_{\text{ref}}; \Theta),\, y\bigr),
\end{equation}
while ensuring that only the parameters in \(\Theta_{\mathcal{T}}\) are updated.

The update rule at iteration \(t\) is formulated as:
\begin{equation}
\Theta^{(t+1)} = \Theta^{(t)} - \eta \left( M \odot \nabla_{\Theta} \mathcal{L}\bigl(f(X_{\text{ref}}; \Theta^{(t)}),\, y\bigr) \right),
\end{equation}
where \(\eta\) is the learning rate and \(\odot\) denotes element-wise multiplication. This ensures that gradients are only applied to the task-relevant parameters while the remaining parameters stay frozen.

By jointly optimizing parameter selection and token refinement, our approach enables a mutual enhancement process, where the refined tokens guide more task-specific parameter updates, and the optimized parameters, in turn, refine the token representations. This co-adaptive learning strategy significantly reduces redundant computations and memory usage, ensuring that ViTs focus on the most task-discriminative features for efficient adaptation.

%% file: sec/4_experiment.tex
\begin{table*}[!t]
    \centering
    \vspace{-0.45cm}
    \resizebox{2.1\columnwidth}{!}{
        \begin{tabular}{c|ccccccc|cccc|cccccccc|cc}
        \toprule
         \multirow{2}{*}{\diagbox[height=7\line]{Method \\ \\ \\}{\\ \\ \\ Dataset}} & \multicolumn{7}{c}{{Natural}}                                       & \multicolumn{4}{c}{{Specialized}}                   & \multicolumn{8}{c}{{Structured}}                                                                                      & \multicolumn{2}{c}{{}}     \\
        \cline{2-22}
         & \rotatebox{90}{CIFAR-100} & \rotatebox{90}{Caltech101} & \rotatebox{90}{DTD} & \rotatebox{90}{Flowers102} & \rotatebox{90}{Pets} & \rotatebox{90}{SVHN} & \rotatebox{90}{Sun397} & \rotatebox{90}{Patch Camelyon } & \rotatebox{90}{EuroSAT} & \rotatebox{90}{Resisc45} & \rotatebox{90}{Retinopathy} & \rotatebox{90}{Clevr/count} & \rotatebox{90}{Clevr/distance} & \rotatebox{90}{DMLab} & \rotatebox{90}{KITTI/distance} & \rotatebox{90}{dSprites/loc} & \rotatebox{90}{dSprites/ori} & \rotatebox{90}{SmallNORB/azi} & \rotatebox{90}{SmallNORB/ele} & \rotatebox{90}{Mean Acc.} & \rotatebox{90}{Params. (\%)} \\
        \midrule
        Full~\cite{jia2022visual}                            & 68.9      & 87.7       & 64.3 & 97.2       & 86.9 & 87.4 & 38.8   & 79.7           & 95.7    & 84.2     & 73.9        & 56.3        & 58.6           & 41.7  & 65.5           & 57.5         & 46.7         & 25.7          & 29.1          & 65.57     & 100.00           \\
        Linear~\cite{jia2022visual}                          & 63.4      & 85.0       & 64.3 & 97.0       & 86.3 & 36.6 & 51.0   & 78.5           & 87.5    & 68.6     & 74.0        & 34.3        & 30.6           & 33.2  & 55.4           & 12.5         & 20.0         & 9.6           & 19.2          & 53.00     & 0.05             \\

        \midrule
        BitFit~\cite{zaken2021bitfit}                            & 72.8      & 87.0       & 59.2 & 97.5       & 85.3 & 59.9 & 51.4   & 78.7           & 91.6    & 72.9     & 69.8        & 61.5        & 55.6           & 32.4  & 55.9           & 66.6         & 40.0         & 15.7          & 25.1          & 62.05     & 0.16             \\

        Adapter~\cite{houlsby2019parameter}                         & 74.1      & 86.1       & 63.2 & 97.7       & 87.0 & 34.6 & 50.8   & 76.3           & 88.0    & 73.1     & 70.5        & 45.7        & 37.4           & 31.2  & 53.2           & 30.3         & 25.4         & 13.8          & 22.1          & 55.82     & 0.31             \\

        AdaptFormer~\cite{chen2022adaptformer}   & 70.8 & 91.2 & 70.5 & 99.1 & 90.9 & 86.6 & 54.8 & 83.0 & 95.8 & 84.4 &{76.3} & 81.9 & \underline{64.3} & 49.3 & 80.3 & 76.3 & 45.7  & 31.7 & 41.1 &72.32 & 0.24\\
        
        LoRA~\cite{hu2021lora}                         & 68.1      & 91.4       & 69.8 & 99.0       & 90.5 & 86.4 & 53.1   & 85.1           & 95.8    & 84.7     & 74.2        & \underline{83.0}        & \textbf{66.9}           & 50.4  & 81.4           & 80.2         & 46.6         & \underline{32.2}          & 41.1          & 72.63     & 0.90             \\

        VPT-Shallow~\cite{jia2022visual}                     & 77.7      & 86.9       & 62.6 & 97.5       & 87.3 & 74.5 & 51.2   & 78.2           & 92.0    & 75.6     & 72.9        & 50.5        & 58.6           & 40.5  & 67.1           & 68.7         & 36.1         & 20.2          & 34.1          & 64.85     & 0.13             \\

        VPT-Deep~\cite{jia2022visual}                        & {78.8}      & 90.8       & 65.8 & 98.0       & 88.3 & 78.1 & 49.6   & 81.8           & \underline{96.1}    & 83.4     & 68.4        & 68.5        & 60.0           & 46.5  & 72.8           & 73.6         & 47.9         & \textbf{32.9}          & 37.8          & 69.43     & 0.70             \\

        SSF~\cite{lian2022scaling}                             & 69.0      & 92.6       & {75.1} & {99.4}       & \underline{91.8} & {90.2} & \underline{52.9}   & {87.4}           & 95.9    & \underline{87.4}     & 75.5        & 75.9        & 62.3           & {53.3}  & 80.6           & 77.3         & {54.9}         & 29.5          & 37.9          & 73.10     & 0.28             \\

        
        GPS~\cite{zhang2024gradient}                      & \underline{81.1}      & \textbf{94.2}       & \textbf{75.8} & \underline{99.4}       & {91.7} & \textbf{91.6} & 52.4   & \underline{87.9}           & \textbf{96.2}    & {86.5}     & \underline{76.5}        & 79.9        & 62.6           & \textbf{55.0}  & \underline{82.4}           & \underline{84.0}         & \underline{55.4}         & 29.7          & \textbf{46.1}          & \underline{75.18}     & 0.25              \\
        \midrule
        \textbf{TR-PTS (Ours)} &\textbf{81.2} & \underline{93.9} & \underline{75.1} &\textbf{99.5} &\textbf{91.9} &\underline{91.0} &\textbf{54.5} &\textbf{88.1} &95.7 &\textbf{87.8} &\textbf{76.6} &\textbf{83.5} &63.2  &\underline{54.8} &\textbf{82.8} &\textbf{87.7} &\textbf{56.9} &31.8 &\textbf{46.1} &\textbf{75.92} &0.34 \\ 
        
        \bottomrule
        \end{tabular}
    }
    \vspace{-0.15cm}
    \caption{\textbf{Performance Comparisons on VTAB-1k with ViT-B/16 Models Pre-trained on ImageNet-21K.}}
    \label{tab:vtab}
    \vspace{-0.2cm}
\end{table*}

\begin{figure*}[!ht]
    \centering
    \vspace{-0.2cm}
    \includegraphics[width=0.95\linewidth]{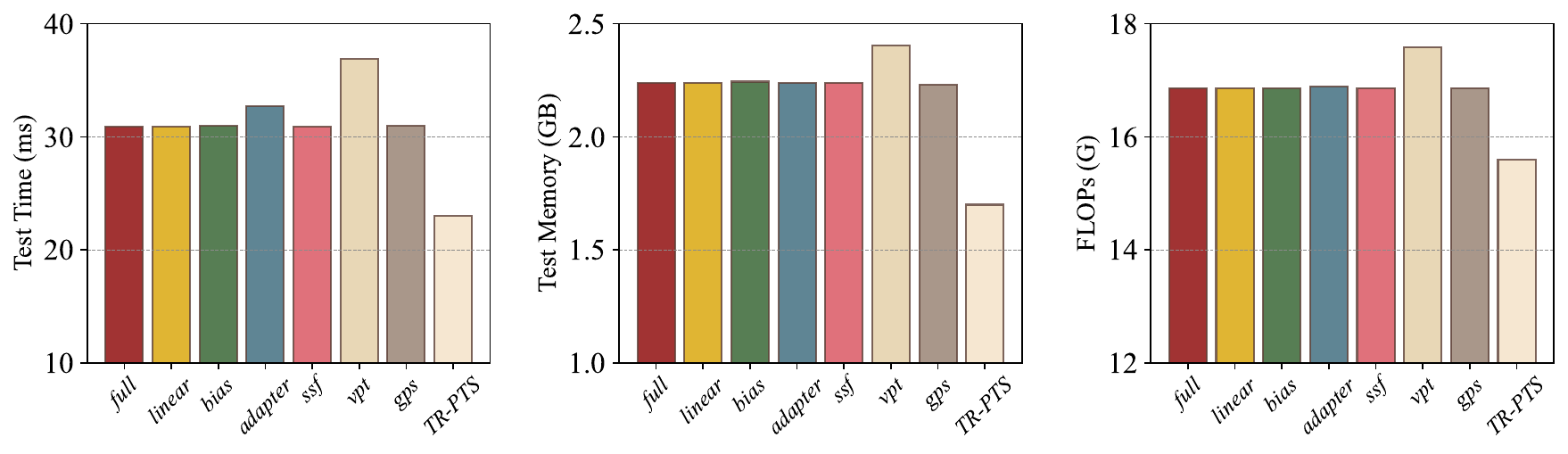}
    \vspace{-0.35cm}
    \caption{\textbf{Comparison of Different Methods in terms of Test Time (ms), Test Memory Usage (GB), and FLOPs (G).} The results show that the TR-PTS method achieves the lowest FLOPs and memory consumption while maintaining competitive test time.}
    \vspace{-0.35cm}
    \label{fig:cost}
\end{figure*}

\begin{table}
	\centering
	\setlength{\tabcolsep}{1pt}
	\scalebox{0.65}{\begin{tabular}{c|c|c|c|c|c|c|c}
			\toprule
			\diagbox{Method}{Dataset}&\makecell[c]{~CUB-200~ \\ -2011} & ~NABirds~ & \makecell[c]{~Oxford~ \\ Flowers}  & \makecell[c]{~Stanford~ \\ Dogs}  & \makecell[c]{~Stanford~ \\ Cars}  & ~~Mean~~ & \makecell[c]{~Params.(\%)} \tabularnewline \midrule
			Full~\cite{jia2022visual}         & 87.3         & 82.7    & 98.8           & 89.4          & 84.5          & 88.54     & 100.00           \\
            Linear~\cite{jia2022visual}         & 85.3         & 75.9    & 97.9           & 86.2          & 51.3          & 79.32     & 0.21             \\
            \midrule
            BitFit~\cite{zaken2021bitfit}           & 88.4         & 84.2    & 98.8           & 91.2          & 79.4          & 88.40     & 0.33             \\
            Adapter~\cite{houlsby2019parameter}       & 87.1         & 84.3    & 98.5           & 89.8          & 68.6          & 85.66     & 0.48             \\
    
            AdaptFormer\cite{chen2022adaptformer} &88.4 &84.7 &99.2 &88.2 &81.9 &88.48 &0.75 \\
                    
            LoRA~\cite{hu2021lora}           & 85.6         & 79.8    & 98.9           & 87.6          & 72.0          & 84.78     & 0.90             \\
            VPT-Shallow~\cite{jia2022visual}    & 86.7         & 78.8    & 98.4           & 90.7          & 68.7          & 84.62     & 0.29             \\
            VPT-Deep~\cite{jia2022visual}       & 88.5         & 84.2    & 99.0           & 90.2          & 83.6          & 89.11     & 0.99             \\
            SSF~\cite{lian2022scaling}            & {89.5}         & {85.7}    & \underline{99.6}           & 89.6          & {89.2}          & {90.72}     & 0.45             \\
            GPS~\cite{zhang2024gradient}     & \underline{89.9}         & \underline{86.7}    & \textbf{99.7}           & \underline{92.2}          & \underline{90.4}          & \underline{91.78}     & 0.77             \\
    
            \midrule
            \textbf{TR-PTS (Ours)} &\textbf{90.0} &\textbf{87.1} & \underline{99.6} &  \textbf{92.4} & \textbf{90.6} & \textbf{91.94} & 0.60
            \\
            \bottomrule
		\end{tabular}
	}
        \vspace{-0.3cm}
        \caption{\textbf{Performance Comparisons on Five FGVC Datasets with ViT-B/16 Models Pre-trained on ImageNet-21K.}}
	\label{table: fgvc}
	\vspace{-0.45cm}
\end{table}

\section{Experiments}
\label{sec:exp}
\subsection{Experimental settings}

\paragraph{Datasets.} We conduct our experiments primarily on a series of datasets categorized into three types as detailed below: 1) \textbf{\textit{{FGVC}}}: Fine-Grained Visual Classification (FGVC) benchmark~\cite{jia2022visual} includes five downstream tasks, which are CUB-200-2011~\cite{wah2011caltech}, NABirds~\cite{van2015building}, Oxford
Flowers~\cite{nilsback2008automated}, Stanford Dogs~\cite{dataset2011novel} and Stanford Cars~\cite{gebru2017fine}. 2) \textbf{\textit{{VTAB-1k}}}: Visual Task Adaptation Benchmark (VTAB)~\cite{zhai2019large} contains of 19 visual classification tasks, which are grouped into three sets: Natural, Specialized, and Structured. Each task in VTAB-1k contains 1000 training example. 

\paragraph{Pre-Trained Backbones.} To ensure a fair comparison, we align with most of efficient fine-tuning methods~\cite{jia2022visual,lian2022scaling,zhang2024gradient} and adopt the plain Vision Transformer~\cite{dosovitskiy2020image}, \textit{i.e.}, ViT-Base (ViT-B/16) as our backbone model and pre-train the model with both supervised method. Specifically, we directly use the ImageNet-21k~\cite{ridnik2021imagenet} supervised pre-trained model.

\paragraph{Baseline Methods.} We compare our TR-PTS  with a variety of fine-tuning protocols that can be mainly categorized into three types~\cite{xin2024parameter}: 1) \textbf{\textit{{Full Fine-tuning}}}: the most commonly used protocol updating all parameters of the whole model during tuning. 2) \textbf{\textit{Partial-based Tuning}}: This kind of method concentrates on updating only a small subset of inherent parameters while maintaining the majority of the model’s parameters unchanged during the adaptation process, including Linear Probing, BitFit~\cite{zaken2021bitfit}, LoRA~\cite{hu2021lora}, SSF~\cite{lian2022scaling}, and GPS~\cite{zhang2024gradient}. 3) \textbf{\textit{Addition-based Tuning}}: This kind of method involves incorporating additional trainable modules or parameters into pre-trained backbones to learn task-specific information, including Adapter~\cite{houlsby2019parameter}, AdaptFormer~\cite{chen2022adaptformer}, VPT~\cite{jia2022visual}.

\vspace{-0.1cm}
\paragraph{Implementation Details.}
We conduct our experiments on two main benchmarks: FGVC and VTAB-1k. For a fair comparison with existing PEFT methods, models are fine-tuned using the Adam optimizer~\cite{kingma2014adam} with a cosine learning rate decay schedule for 100 epochs. All experiments are implemented using the PyTorch framework~\cite{paszke2019pytorch}.
\vspace{-0.1cm}
\subsection{Main Properties and Analysis}
\label{sec_mainexp}
We conduct a comprehensive evaluation on two benchmarks, VTAB and FGVC, which together consist of 24 diverse datasets. In our experiments, we compare our proposed approach with leading fine tuning protocols based on Top-1 accuracy and the percentage of fine-tuned parameters. In addition, we rigorously assess both computational and storage costs to demonstrate the efficiency of our approach, thereby validating its practical advantages in terms of both performance and resource utilization.

\begin{figure*}[!t]
    \centering
    \vspace{-0.5cm}
    \begin{subfigure}{0.24\textwidth}
        \includegraphics[width=\textwidth]{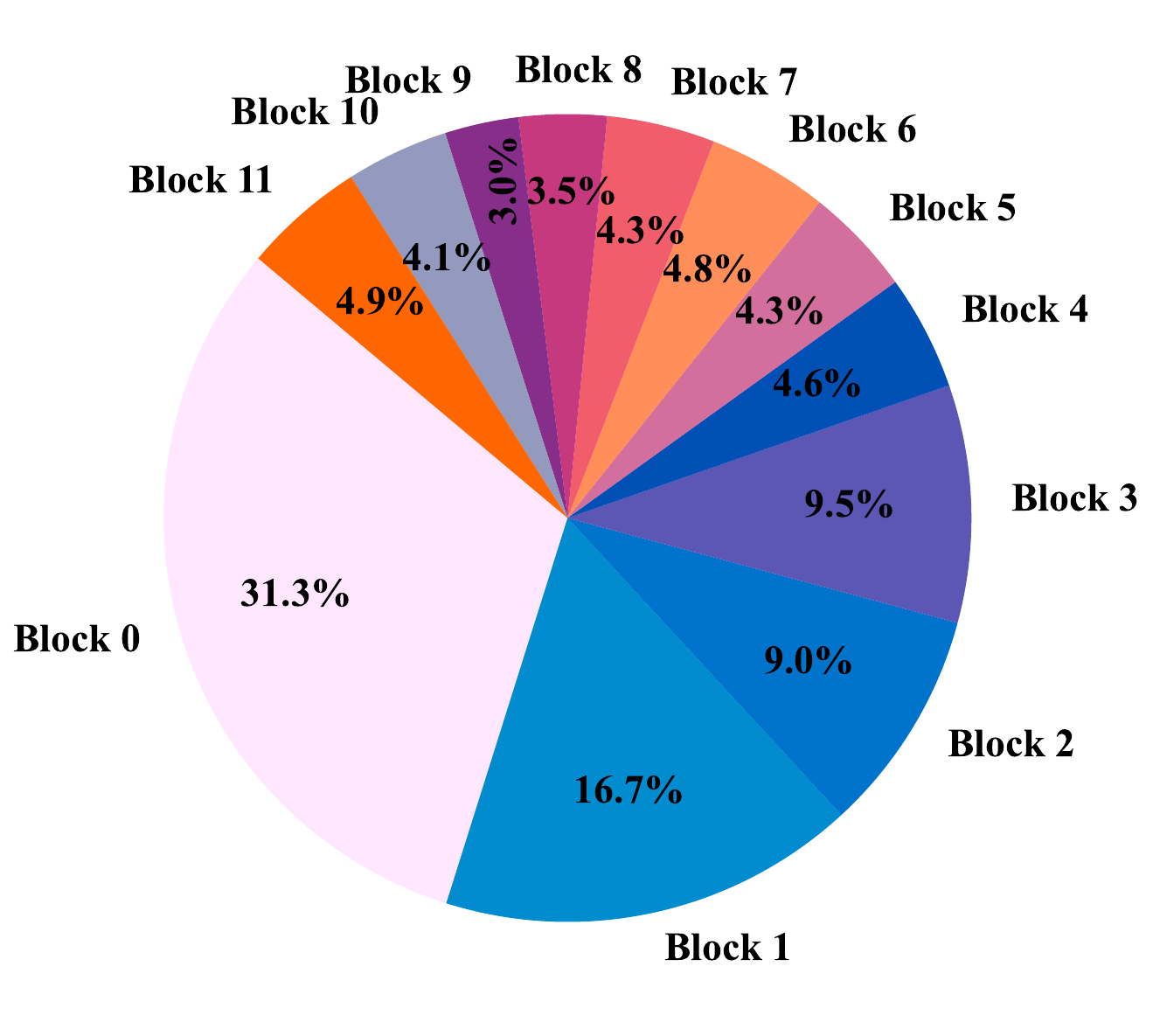}
        \caption{Sun397}
    \end{subfigure}
    \begin{subfigure}{0.24\textwidth}
        \includegraphics[width=\textwidth]{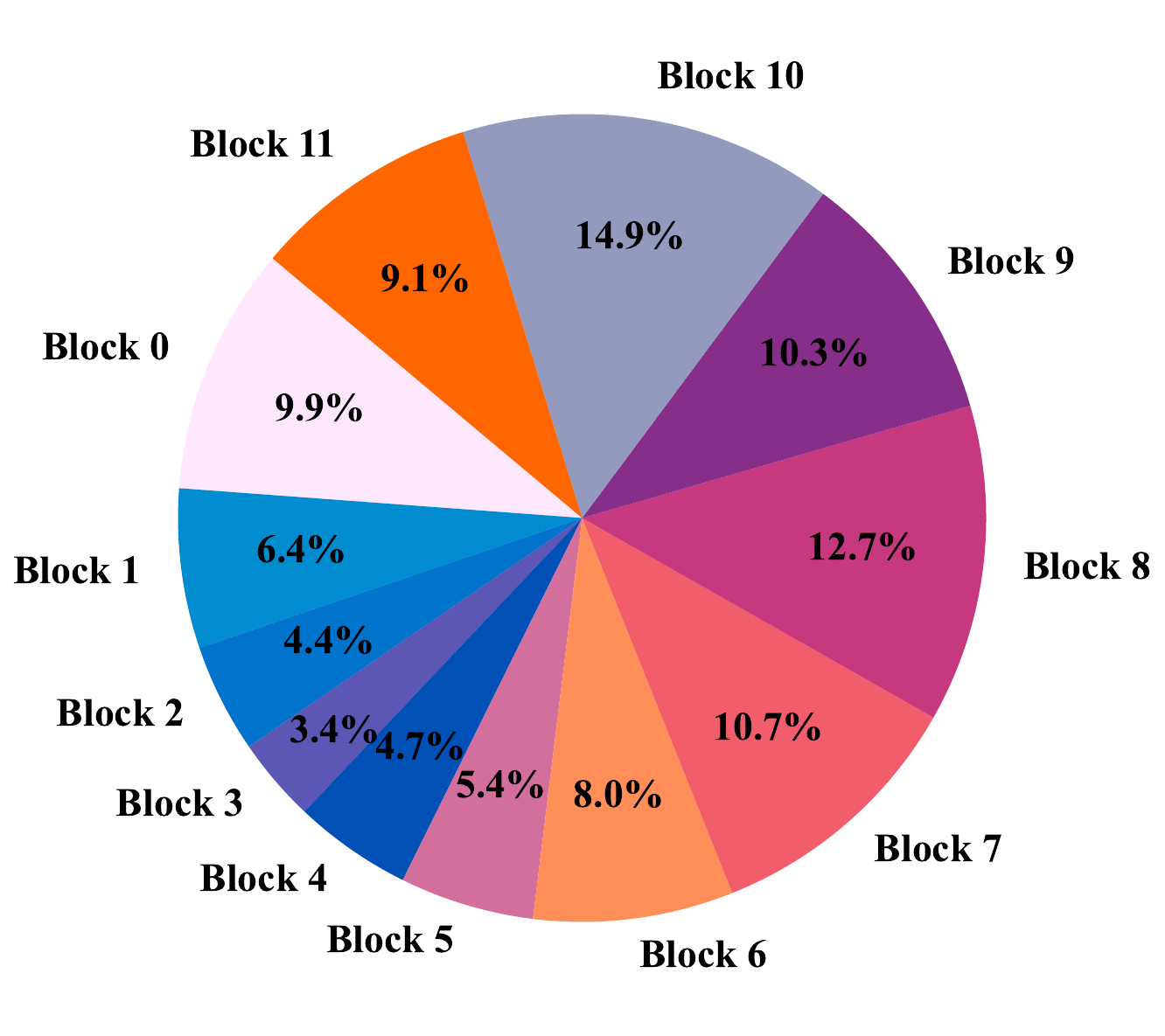}
        \caption{Flower102}
    \end{subfigure}
    \begin{subfigure}{0.24\textwidth}
        \includegraphics[width=\textwidth]{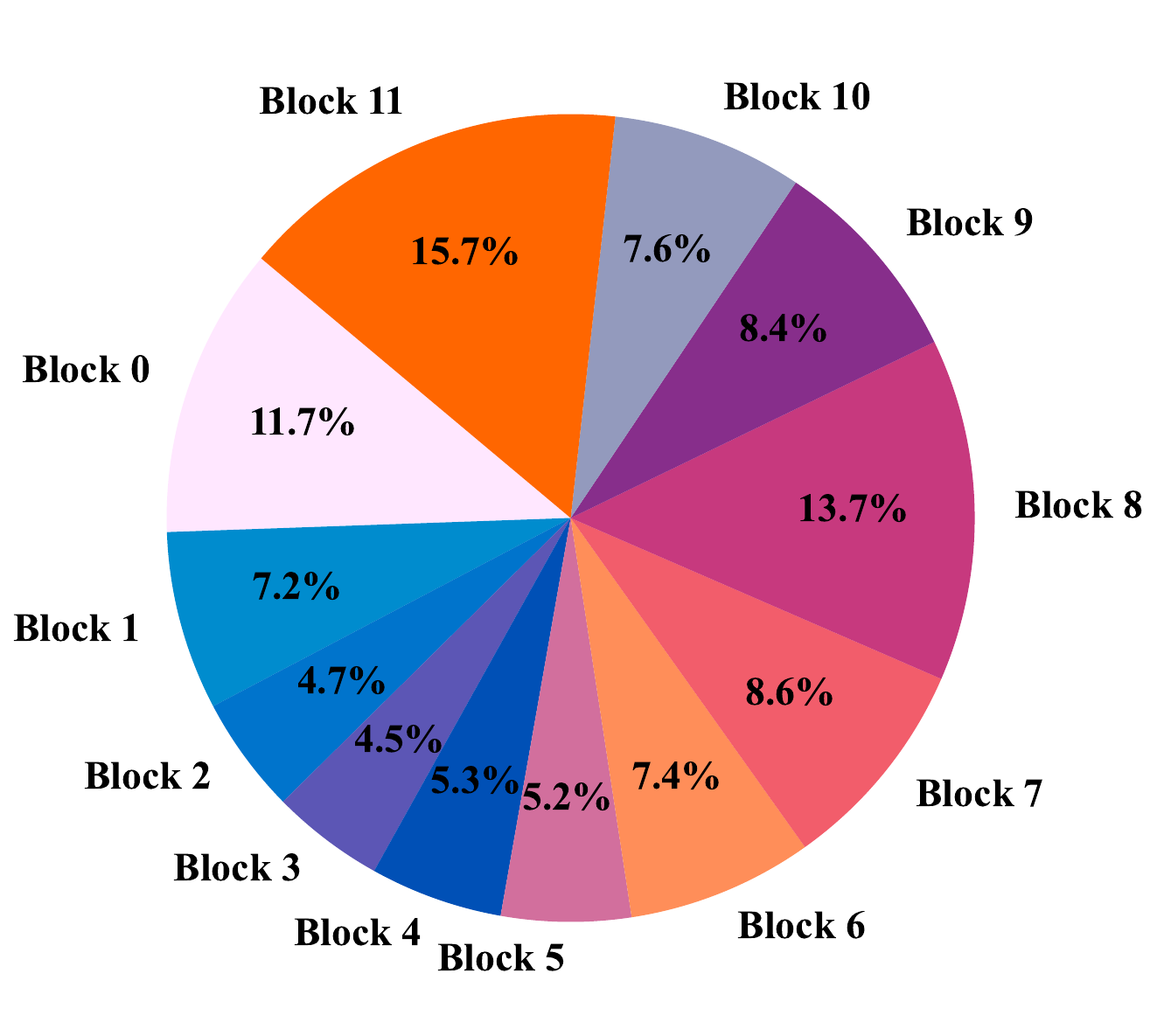}
        \caption{Dsprites/loc}
    \end{subfigure}
    \begin{subfigure}{0.24\textwidth}
        \includegraphics[width=\textwidth]{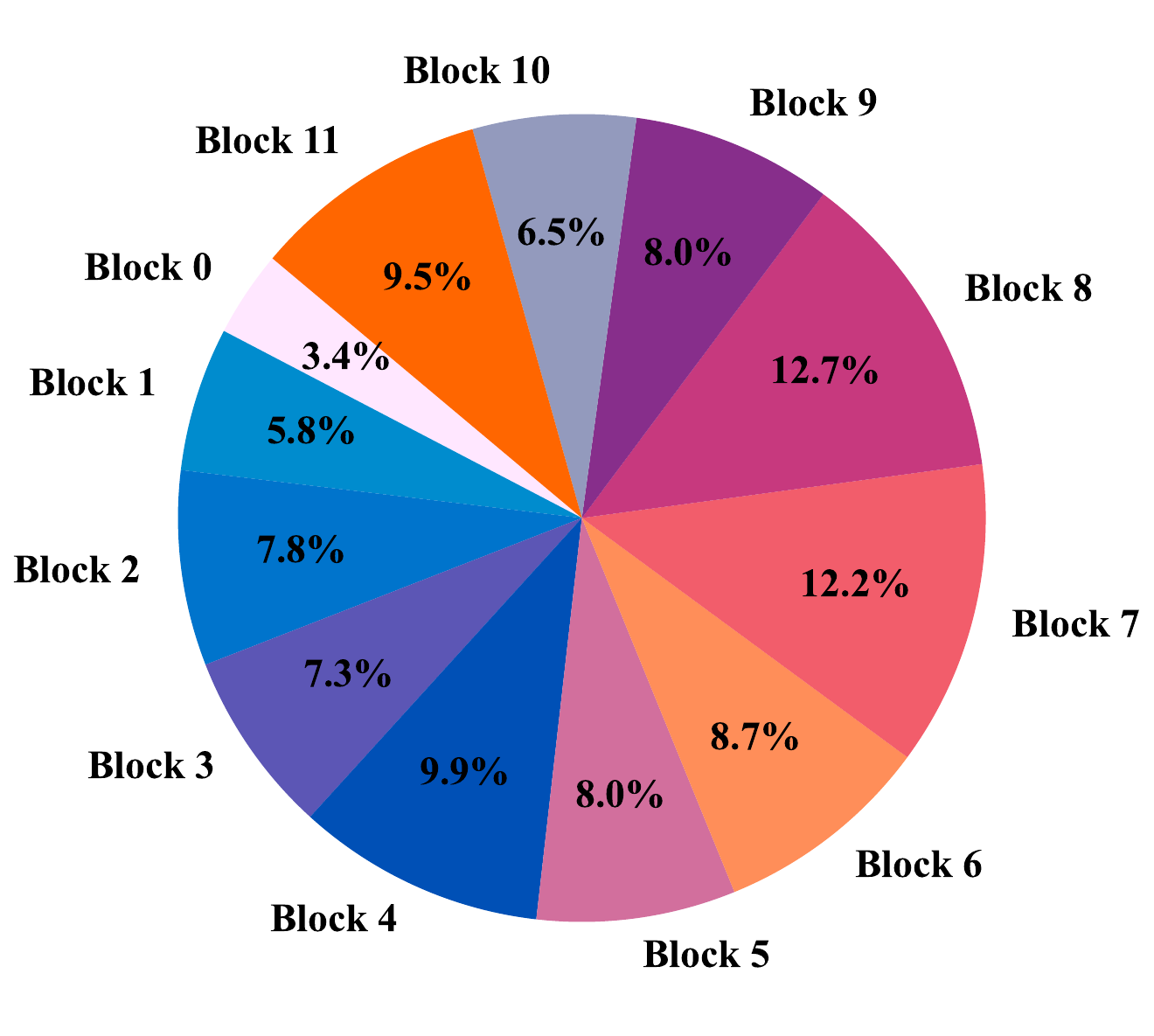}
        \caption{Patch Camelyon}
    \end{subfigure}
    \vspace{-0.1cm}
    \caption{\textbf{Differernt Datasets Top 1\% FIM Parameter Distribution.}}
    \vspace{-0.2cm}
    \label{fig:dataset_distribution}
\end{figure*}
\vspace{-0.1cm}
\paragraph{Comparisons on VTAB-1K.}
For the VTAB benchmark, our TR-PTS framework achieves substantial improvements over existing fine-tuning methods. As shown in Table~\ref{tab:vtab}, it attains an average top-1 accuracy of 75.92\%, exceeding full fine-tuning by 10.35\% and surpassing GPS by 0.74\%, demonstrating superior performance on 13 subtasks and state-of-the-art results on 11. Notably, TR-PTS fine-tunes only 0.34\% of model parameters, significantly reducing computational cost. The superior performance of TR-PTS stems from its task-relevant token and parameter selection strategies: layer-wise allocation refines only the most critical parameters, improving accuracy while minimizing parameter updates, whereas token selection enhances efficiency by prioritizing task-critical information. By selectively merging less informative tokens, our method prevents redundancy while preserving essential representations. These combined strategies enable TR-PTS to achieve both efficiency and robustness, making it highly effective for the VTAB benchmark.
\vspace{-0.1cm}
\paragraph{Comparisons on FGVC.}
Table~\ref{table: fgvc} shows that TR-PTS achieves the state of art performance on FGVC datasets, with an average accuracy of 91.94\%, slightly surpassing GPS (91.78\%) and and outperforming full fine-tuning by 3.40\%. While the accuracy gains are marginal, fine-grained classification presents inherent challenges, with subtle inter-class differences limiting performance improvements. 
What distinguishes TR-PTS is its efficiency. Unlike other methods, our token selection strategy effectively reduces computational cost, as quantified by FLOPs (see Figure~\ref{fig:cost}). These results demonstrate that TR-PTS not only delivers competitive accuracy in a demanding fine-grained classification setting but also significantly enhances computational efficiency.

\paragraph{Computational Cost.}
In Figure~\ref{fig:cost}, we evaluate TR-PTS against various PEFT methods in terms of computational efficiency, with all experiments conducted on a single NVIDIA A100 GPU under consistent settings. Specifically, all methods are evaluated using a batch size of 32, input resolution of $224 \times 224$, and a ViT-B backbone. FLOPs are measured analytically, inference time is averaged over 500 forward passes, and memory usage is recorded after a single pass. TR-PTS achieves the shortest inference time, primarily due to its task-relevant token selection mechanism, which removes redundant tokens and accelerates processing. In terms of memory usage, TR-PTS consumes the least among all methods, unlike VPT, which introduces additional modules that increase runtime memory requirements. Regarding FLOPs, a key indicator of computational complexity, TR-PTS reports the lowest count by dynamically selecting and merging tokens, thereby reducing unnecessary operations. By comparison, GPS lacks structured token selection and incurs higher FLOPs. These results demonstrate that TR-PTS effectively enhances both token and parameter efficiency, reducing computational cost while preserving strong task performance.

\subsection{Ablation Studies}
\paragraph{Components Effectiveness.}
We analyze the contribution of each component in our framework: Task-Relevant Parameter Selection and Task-Relevant Token Selection. Table~\ref{tab:ablation_framework} presents the impact of integrating these components. To ensure a comprehensive evaluation, we select one task from each of the three major VTAB categories and use linear fine-tuning as the baseline.
Introducing TR-PTS significantly improves performance, achieving gains of 75.2\%, 2.5\%, and 3.5\% across datasets. Applying token selection at an optimal layer, without additional parameter selection or fine-tuning, yields accuracy improvements of 2.3\%, 1.8\%, and 0.2\%, respectively. Conversely, parameter selection alone, without token selection, leads to increases of 72.6\%, 2.4\%, and 3.2\%. The combination of both components in TR-PTS consistently enhances model performance. These results highlight the necessity of selecting both task-relevant tokens and parameters, demonstrating their complementary roles in optimizing fine-tuning for diverse downstream tasks.

\begin{table}[t]
    \centering 
    \resizebox{0.45\textwidth}{!}{
    \LARGE  
    \begin{tabular}{ccccc}
    \toprule
     \multicolumn{2}{c}{TR-PTS}          & \multicolumn{3}{c}{VTAB-1k} \\
    \cmidrule(lr){1-2}\cmidrule(lr){3-5}
    Token Selection & Parameter Selecetion           & dSprites/loc & Flower102 &Sun397       \\ \midrule
    \usym{2613} & \usym{2613} & 12.5 & 97.0 & 51.0 \\
    \checkmark  & \usym{2613} & 14.8 & 98.8 & 51.2 \\
    \usym{2613}  & \checkmark & 85.1 & 99.4 & 54.2 \\
    \checkmark  & \checkmark & \textbf{87.7} & \textbf{99.5} & \textbf{54.5} \\
    \bottomrule
    \end{tabular}
}
\vspace{-0.1cm}
\caption{
\textbf{Ablation Study on the impact of Task-Relevant Token Selection and Task-Relevant Parameter Selection.}}
\vspace{-0.5cm}
\label{tab:ablation_framework}
\end{table}

\begin{figure}[t]
    \centering
    \includegraphics[width=0.9\linewidth]{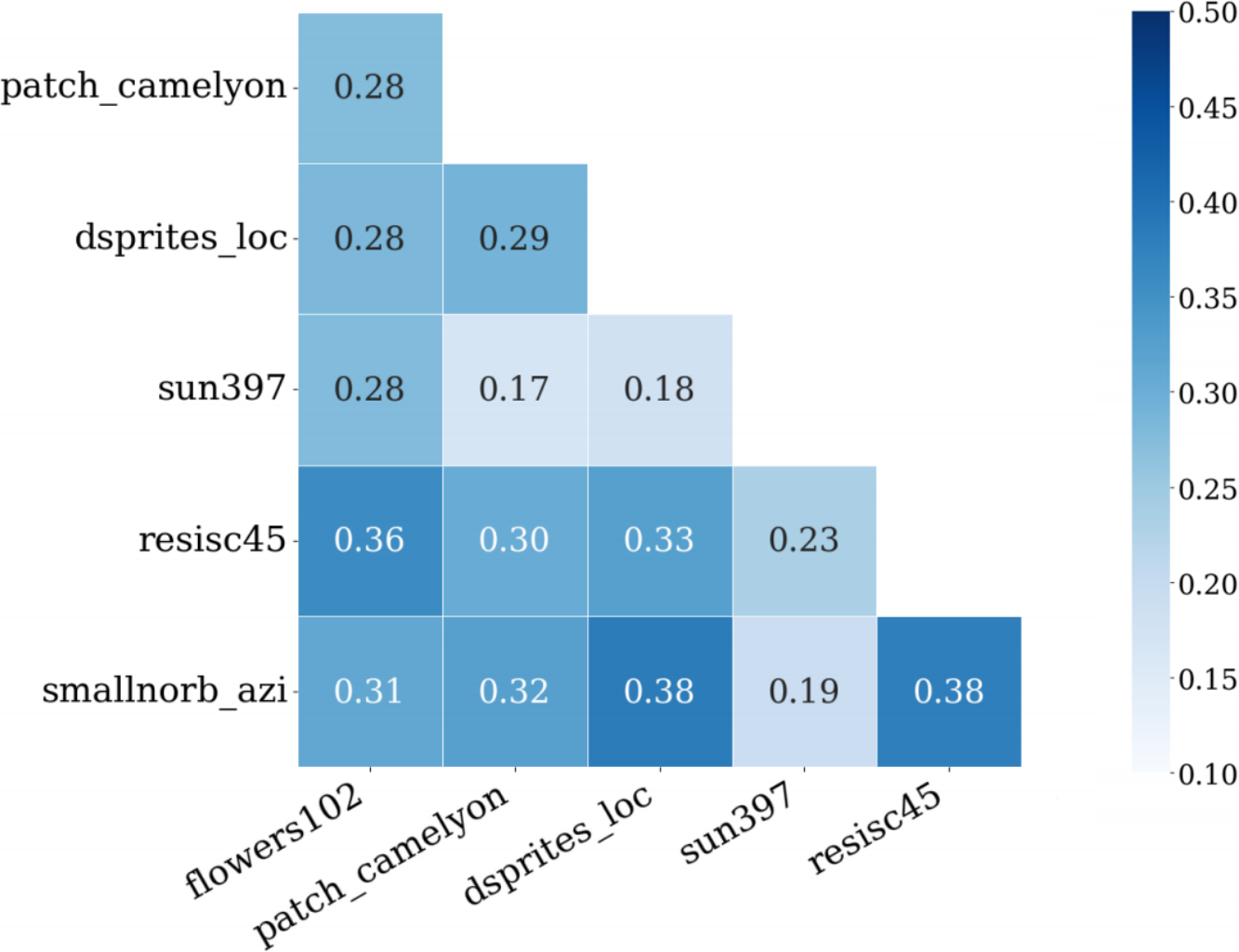}
    \vspace{-0.25cm}
    \caption{\textbf{Visualization of the Overlapping Rate Among Task-Driven Parameter Sets.} This heatmap shows the overlap between parameter sets across tasks. Darker shades indicate higher similarity, while lighter shades highlight task-specific differences.}
    \vspace{-0.5cm}
    \label{fig:hm_less}
\end{figure}

\paragraph{Analysis of Task-Relevant Parameter Selection.}
To assess the impact of our task-relevant parameter selection strategy, we analyze the layer-wise distribution of critical parameters and the overlap between task-specific parameter sets.
\begin{itemize}
    \item \textbf{Layer-Wise Distribution of Critical Parameters: }The parameter distribution across datasets reflects task relevance, varying by dataset needs. As shown in Figure~\ref{fig:dataset_distribution}, Flower102 concentrates parameters in Blocks 8 and 10, relying more on high-level feature extraction, while lower layers contribute less. In contrast, Patch/Camelyon has a uniform distribution, indicating equal reliance on all blocks. Sun397, however, is dominated by Block 0, suggesting a greater dependency on low-level features. These results confirm that task-aware parameter selection is crucial, as different datasets prioritize different layers.

    \item \textbf{Overlap Between Task-Specific Parameter Sets:} As shown in Figure~\ref{fig:hm_less}, similar tasks exhibit greater overlap, such as Resisc45 and Smallnorb/Azi (0.38), indicating shared task-relevant parameters. In contrast, Sun397 shows minimal overlap with Patch/Camelyon (0.17) and Dsprites/Loc (0.18), suggesting distinct parameter selection. Moderate overlap between Dsprites/Loc and Patch/Camelyon (0.28–0.32) implies partial parameter sharing while maintaining task specificity. Overall, the low parameter overlap validates the effectiveness of our adaptive selection strategy over uniform allocation.
\end{itemize}
 
\begin{figure}[t]
\centering
\vspace{-0.35cm}
\includegraphics[width=1\linewidth]{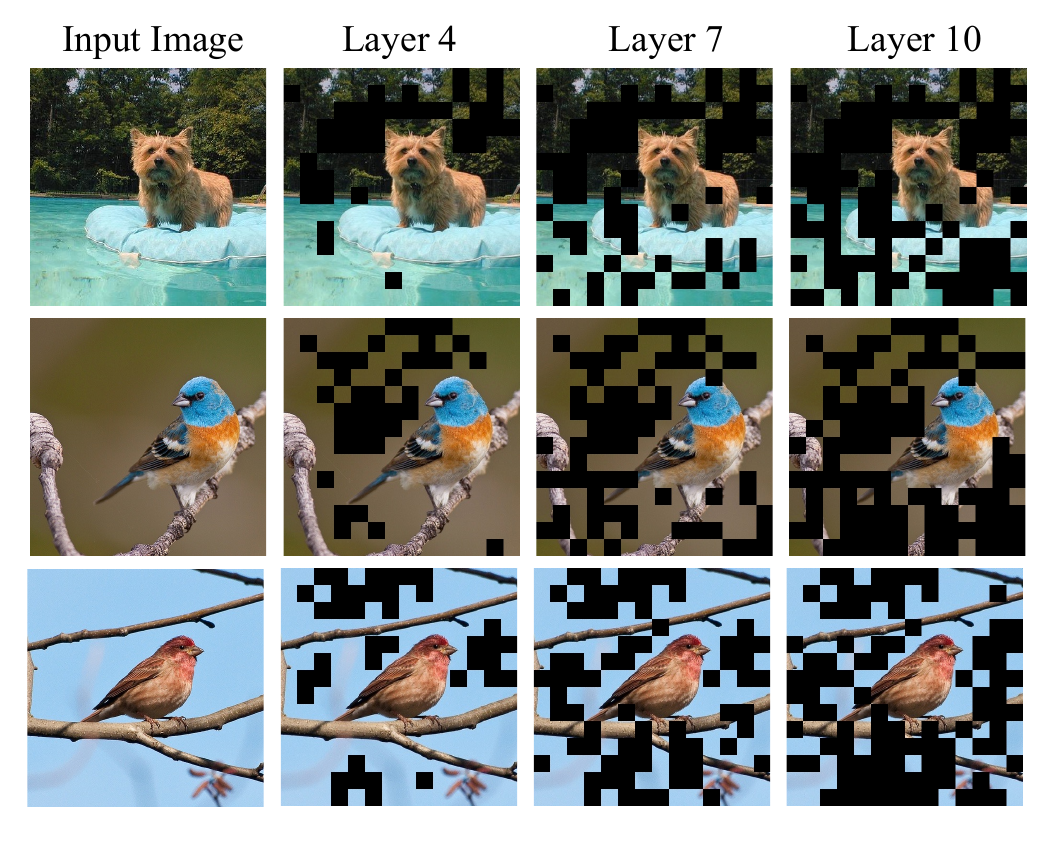}
\vspace{-0.75cm}
\caption{\textbf{Visualization of Task-Relevant Tokens Selected by TR-PTS Using ViT-B/16 with 12 Layers.} As the number of layers increases, our method increasingly focuses on task-relevant tokens.}
\label{fig:token_merge}
\vspace{-0.2cm}
\end{figure}

\paragraph{Explore the Intrinsic Correlation Between Parameter Density and Token Redundancy.}
By analyzing token selection and parameter distribution, we observe a strong intrinsic correlation between token redundancy and parameter sparsity. In particular, layers with fewer task-relevant parameters tend to encode less informative tokens, making them better candidates for token reduction. As shown in Table~\ref{table:merge_position}, applying token selection in layers with dense parameter updates, such as in Flower102 and Camelyon, often leads to performance drops. This indicates that pruning in highly task-relevant layers can disrupt essential task-specific information. Moreover, random token selection across layers results in inconsistent performance, further emphasizing the importance of a task-aware strategy. Motivated by this finding, we design a joint selection strategy that applies token reduction primarily in sparse parameter layers—those with fewer selected task-relevant connections. This “sparse insertion” method ensures that token selection minimizes interference with critical computation while still reducing redundancy. As a result, TR-PTS achieves a favorable balance between efficiency and accuracy, outperforming both dense and random strategies.


\begin{table}[t]
    \setlength{\tabcolsep}{2pt} 
    \renewcommand{\arraystretch}{1.1} 
    \centering
    \small
    \resizebox{0.45\textwidth}{!}{
    \begin{tabular}{l c|cccc}
        \toprule
        & Selection Ratio & Sun397 & Flower102 & Loc & Camelyon \\
        \midrule
        \multirow{2}{*}{Dense} & \textit{0.95} & 53.5 & 99.3 & 85.2 & 87.3 \\
                                       & \textit{0.8}  &  52.7 &99.1 & 86.9 & 87.4 \\
        \midrule
        \multirow{2}{*}{Random} & \textit{0.95} &  54.0 &99.3   & 85.9 & 87.9 \\
                                & \textit{0.8}  &  52.5  &99.2 & 85.5 & 86.1 \\
        \midrule
        \multirow{2}{*}{Sparse} & \textit{0.95} &  \textbf{54.5} &\textbf{99.4} & \textbf{87.7} & \textbf{88.1} \\
                                         & \textit{0.8}  &  52.9 &99.1 & 86.0 & 87.2 \\
        \bottomrule
    \end{tabular}
    }
    \caption{\textbf{Comparison of Token Selection Positions.} “Dense” refers to layers rich in task-relevant parameters, where pruning may harm performance. “Sparse” applies selection in layers with fewer task-relevant parameters, making pruning more effective. “Random” selects tokens without a structured strategy.}
    \vspace{-0.1cm}
\label{table:merge_position}
\end{table}
\paragraph{Token Selection Visualization.}
To evaluate the effectiveness of our proposed Task-Relevant Token Selection strategy, we conduct a Token Selection Visualization experiment using ViT-B/16 with 12 layers. This experiment is performed on two fine-grained datasets, CUB-200-2011 and Stanford Dogs, with a token select rate of 0.8. We visualize the token selection process at different layers, specifically the 4th, 7th, and 10th layers, all within a single forward pass of the network. The results show that in the early layers (e.g., Layer 4), token retention is relatively widespread, capturing local features. In deeper layers (e.g., Layer 10), the model gradually focuses on foreground objects, such as birds in CUB-200-2011 and dogs in Stanford Dogs, effectively eliminating background noise. As the network deepens, more tokens are merged or refined, reducing computational redundancy while allowing the model to concentrate on the most critical image regions.


%% file: sec/5_conclusion.tex
\section{Conclusion}
This work enhances the efficiency of adapting pre-trained Vision Transformers (ViT) during fine-tuning and inference. We propose Task-Relevant Parameter and Token Selection (TR-PTS), which improves efficiency from two perspectives. On the parameter side, it leverages the Fisher Information Matrix (FIM) to fine-tune only the most task-relevant parameters, reducing trainable parameters while preserving adaptation. On the token side, TR-PTS dynamically selects and merges tokens based on $[CLS]$ attention scores, retaining only the most informative ones to reduce redundancy and lower computational overhead. TR-PTS maintains strong task performance with reduced cost. Future work will extend it to segmentation, detection, and adaptive token selection across Transformer layers to enhance efficiency and generalization.
\section{Acknowledge}
The work was supported by the National Natural Science Foundation of China under Grant 62301310.

%% file: main.bbl
\begin{thebibliography}{39}
\providecommand{\natexlab}[1]{#1}
\providecommand{\url}[1]{\texttt{#1}}
\expandafter\ifx\csname urlstyle\endcsname\relax
  \providecommand{\doi}[1]{doi: #1}\else
  \providecommand{\doi}{doi: \begingroup \urlstyle{rm}\Url}\fi

\bibitem[Bolya et~al.(2023)Bolya, Fu, Dai, Zhang, Feichtenhofer, and
  Hoffman]{bolya2023tokenmergingvitfaster}
Daniel Bolya, Cheng-Yang Fu, Xiaoliang Dai, Peizhao Zhang, Christoph
  Feichtenhofer, and Judy Hoffman.
\newblock Token merging: Your vit but faster, 2023.

\bibitem[Brown et~al.(2020)Brown, Mann, Ryder, Subbiah, Kaplan, Dhariwal,
  Neelakantan, Shyam, Sastry, Askell, et~al.]{brown2020language}
Tom Brown, Benjamin Mann, Nick Ryder, Melanie Subbiah, Jared~D Kaplan, Prafulla
  Dhariwal, Arvind Neelakantan, Pranav Shyam, Girish Sastry, Amanda Askell,
  et~al.
\newblock Language models are few-shot learners.
\newblock In \emph{Advances in Neural Information Processing Systems
  (NeurIPS)}, 2020.

\bibitem[Chen et~al.(2022)Chen, Ge, Tong, Wang, Song, Wang, and
  Luo]{chen2022adaptformer}
Shoufa Chen, Chongjian Ge, Zhan Tong, Jiangliu Wang, Yibing Song, Jue Wang, and
  Ping Luo.
\newblock Adaptformer: Adapting vision transformers for scalable visual
  recognition.
\newblock \emph{Advances in Neural Information Processing Systems (NeurIPS)},
  2022.

\bibitem[Dataset(2011)]{dataset2011novel}
E Dataset.
\newblock Novel datasets for fine-grained image categorization.
\newblock In \emph{Proceedings of the IEEE Conference on Computer Vision and
  Pattern Recognition Workshops (CVPR Workshops)}, 2011.

\bibitem[Devlin et~al.(2018)Devlin, Chang, Lee, and Toutanova]{devlin2018bert}
Jacob Devlin, Ming-Wei Chang, Kenton Lee, and Kristina Toutanova.
\newblock Bert: Pre-training of deep bidirectional transformers for language
  understanding.
\newblock \emph{Proceedings of the Annual Conference of the North American
  Chapter of the Association for Computational Linguistics (NAACL)}, 2018.

\bibitem[Dosovitskiy et~al.(2020)Dosovitskiy, Beyer, Kolesnikov, Weissenborn,
  Zhai, Unterthiner, Dehghani, Minderer, Heigold, Gelly,
  et~al.]{dosovitskiy2020image}
Alexey Dosovitskiy, Lucas Beyer, Alexander Kolesnikov, Dirk Weissenborn,
  Xiaohua Zhai, Thomas Unterthiner, Mostafa Dehghani, Matthias Minderer, Georg
  Heigold, Sylvain Gelly, et~al.
\newblock An image is worth 16x16 words: Transformers for image recognition at
  scale.
\newblock In \emph{Proceedings of the International Conference on Learning
  Representations (ICLR)}, 2020.

\bibitem[Dosovitskiy et~al.(2021)Dosovitskiy, Beyer, Kolesnikov, Weissenborn,
  Zhai, Unterthiner, Dehghani, Minderer, Heigold, Gelly, Uszkoreit, and
  Houlsby]{dosovitskiy2020vit}
Alexey Dosovitskiy, Lucas Beyer, Alexander Kolesnikov, Dirk Weissenborn,
  Xiaohua Zhai, Thomas Unterthiner, Mostafa Dehghani, Matthias Minderer, Georg
  Heigold, Sylvain Gelly, Jakob Uszkoreit, and Neil Houlsby.
\newblock An image is worth 16x16 words: Transformers for image recognition at
  scale.
\newblock \emph{Proceedings of the International Conference on Learning
  Representations (ICLR)}, 2021.

\bibitem[Du et~al.(2025)Du, Luo, Xin, Chen, Feng, Zhang, and Wang]{du2025multi}
Yuntao Du, Siqi Luo, Yi Xin, Mingcai Chen, Shuai Feng, Mujie Zhang, and
  Chonngjun Wang.
\newblock Multi-source fully test-time adaptation.
\newblock \emph{Neural Networks}, 181:\penalty0 106661, 2025.

\bibitem[Fayyaz et~al.(2022)Fayyaz, Koohpayegani, Jafari, Sengupta, Joze,
  Sommerlade, Pirsiavash, and Gall]{fayyaz2022adaptive}
Mohsen Fayyaz, Soroush~Abbasi Koohpayegani, Farnoush~Rezaei Jafari, Sunando
  Sengupta, Hamid Reza~Vaezi Joze, Eric Sommerlade, Hamed Pirsiavash, and
  J{\"u}rgen Gall.
\newblock Adaptive token sampling for efficient vision transformers.
\newblock In \emph{European Conference on Computer Vision}, pages 396--414.
  Springer, 2022.

\bibitem[Gebru et~al.(2017)Gebru, Krause, Wang, Chen, Deng, and
  Fei-Fei]{gebru2017fine}
Timnit Gebru, Jonathan Krause, Yilun Wang, Duyun Chen, Jia Deng, and Li
  Fei-Fei.
\newblock Fine-grained car detection for visual census estimation.
\newblock In \emph{Proceedings of the AAAI Conference on Artificial
  Intelligence (AAAI)}, 2017.

\bibitem[Houlsby et~al.(2019)Houlsby, Giurgiu, Jastrzebski, Morrone,
  De~Laroussilhe, Gesmundo, Attariyan, and Gelly]{houlsby2019parameter}
Neil Houlsby, Andrei Giurgiu, Stanislaw Jastrzebski, Bruna Morrone, Quentin
  De~Laroussilhe, Andrea Gesmundo, Mona Attariyan, and Sylvain Gelly.
\newblock Parameter-efficient transfer learning for nlp.
\newblock In \emph{Proceedings of the International Conference on Machine
  Learning (ICML)}, 2019.

\bibitem[Hu et~al.(2022{\natexlab{a}})Hu, Shen, Wallis, Allen-Zhu, Li, Wang,
  Wang, and Chen]{hu2021lora}
Edward~J Hu, Yelong Shen, Phillip Wallis, Zeyuan Allen-Zhu, Yuanzhi Li, Shean
  Wang, Lu Wang, and Weizhu Chen.
\newblock Lora: Low-rank adaptation of large language models.
\newblock \emph{Proceedings of the International Conference on Learning
  Representations (ICLR)}, 2022{\natexlab{a}}.

\bibitem[Hu et~al.(2022{\natexlab{b}})Hu, Shen, Wallis, Allen-Zhu, Li, Wang,
  Wang, Chen, et~al.]{hu2022lora}
Edward~J Hu, Yelong Shen, Phillip Wallis, Zeyuan Allen-Zhu, Yuanzhi Li, Shean
  Wang, Lu Wang, Weizhu Chen, et~al.
\newblock Lora: Low-rank adaptation of large language models.
\newblock \emph{ICLR}, 1\penalty0 (2):\penalty0 3, 2022{\natexlab{b}}.

\bibitem[Jia et~al.(2022)Jia, Tang, Chen, Cardie, Belongie, Hariharan, and
  Lim]{jia2022visual}
Menglin Jia, Luming Tang, Bor-Chun Chen, Claire Cardie, Serge Belongie, Bharath
  Hariharan, and Ser-Nam Lim.
\newblock Visual prompt tuning.
\newblock In \emph{Proceedings of the European Conference on Computer Vision
  (ECCV)}, 2022.

\bibitem[Kingma and Ba(2014)]{kingma2014adam}
Diederik~P Kingma and Jimmy Ba.
\newblock Adam: A method for stochastic optimization.
\newblock \emph{arXiv preprint arXiv:1412.6980}, 2014.

\bibitem[Kirkpatrick et~al.(2017)Kirkpatrick, Pascanu, Rabinowitz, Veness,
  Desjardins, Rusu, Milan, Quan, Ramalho, Grabska-Barwinska,
  et~al.]{kirkpatrick2017overcoming}
James Kirkpatrick, Razvan Pascanu, Neil Rabinowitz, Joel Veness, Guillaume
  Desjardins, Andrei~A Rusu, Kieran Milan, John Quan, Tiago Ramalho, Agnieszka
  Grabska-Barwinska, et~al.
\newblock Overcoming catastrophic forgetting in neural networks.
\newblock \emph{Proceedings of the national academy of sciences}, 114\penalty0
  (13):\penalty0 3521--3526, 2017.

\bibitem[Lian et~al.(2022)Lian, Zhou, Feng, and Wang]{lian2022scaling}
Dongze Lian, Daquan Zhou, Jiashi Feng, and Xinchao Wang.
\newblock Scaling \& shifting your features: A new baseline for efficient model
  tuning.
\newblock In \emph{Advances in Neural Information Processing Systems
  (NeurIPS)}, 2022.

\bibitem[Liu et~al.(2024)Liu, Liu, Huang, Shi, Xu, Xin, Yin, and
  Liu]{liu2024sparse}
Ting Liu, Xuyang Liu, Siteng Huang, Liangtao Shi, Zunnan Xu, Yi Xin, Quanjun
  Yin, and Xiaohong Liu.
\newblock Sparse-tuning: Adapting vision transformers with efficient
  fine-tuning and inference.
\newblock \emph{arXiv preprint arXiv:2405.14700}, 2024.

\bibitem[Liu et~al.(2025)Liu, Liu, Huang, Xin, Hu, Qin, Wang, Wu, and
  Chen]{liu2025m2ist}
Xuyang Liu, Ting Liu, Siteng Huang, Yi Xin, Yue Hu, Long Qin, Donglin Wang,
  Yuanyuan Wu, and Honggang Chen.
\newblock M2ist: Multi-modal interactive side-tuning for efficient referring
  expression comprehension.
\newblock \emph{IEEE Transactions on Circuits and Systems for Video Technology
  (TCSVT)}, 2025.

\bibitem[Luo et~al.(2024)Luo, Xin, Du, Wan, Tan, Zhai, and
  Liu]{luo2024enhancing}
Siqi Luo, Yi Xin, Yuntao Du, Zhongwei Wan, Tao Tan, Guangtao Zhai, and Xiaohong
  Liu.
\newblock Enhancing test time adaptation with few-shot guidance.
\newblock \emph{arXiv preprint arXiv:2409.01341}, 2024.

\bibitem[Nilsback and Zisserman(2008)]{nilsback2008automated}
Maria-Elena Nilsback and Andrew Zisserman.
\newblock Automated flower classification over a large number of classes.
\newblock In \emph{2008 Sixth Indian conference on computer vision, graphics \&
  image processing}, 2008.

\bibitem[OpenAI(2023)]{openai2023gpt4}
OpenAI.
\newblock Gpt-4 technical report.
\newblock \emph{arXiv preprint arXiv:2304.10592}, 2023.

\bibitem[Paszke et~al.(2019)Paszke, Gross, Massa, Lerer, Bradbury, Chanan,
  Killeen, Lin, Gimelshein, Antiga, et~al.]{paszke2019pytorch}
Adam Paszke, Sam Gross, Francisco Massa, Adam Lerer, James Bradbury, Gregory
  Chanan, Trevor Killeen, Zeming Lin, Natalia Gimelshein, Luca Antiga, et~al.
\newblock Pytorch: An imperative style, high-performance deep learning library.
\newblock \emph{Advances in neural information processing systems}, 32, 2019.

\bibitem[Qin et~al.(2025)Qin, Zhuo, Xin, Du, Li, Fu, Lu, Yuan, Li, Liu,
  et~al.]{qin2025lumina}
Qi Qin, Le Zhuo, Yi Xin, Ruoyi Du, Zhen Li, Bin Fu, Yiting Lu, Jiakang Yuan,
  Xinyue Li, Dongyang Liu, et~al.
\newblock Lumina-image 2.0: A unified and efficient image generative framework.
\newblock \emph{arXiv preprint arXiv:2503.21758}, 2025.

\bibitem[Rao et~al.(2021)Rao, Zhao, Liu, Lu, Zhou, and
  Hsieh]{rao2021dynamicvit}
Yongming Rao, Wenliang Zhao, Benlin Liu, Jiwen Lu, Jie Zhou, and Cho-Jui Hsieh.
\newblock Dynamicvit: Efficient vision transformers with dynamic token
  sparsification.
\newblock \emph{Advances in Neural Information Processing Systems (NeurIPS)},
  2021.

\bibitem[Ridnik et~al.(2021)Ridnik, Ben-Baruch, Noy, and
  Zelnik-Manor]{ridnik2021imagenet}
Tal Ridnik, Emanuel Ben-Baruch, Asaf Noy, and Lihi Zelnik-Manor.
\newblock Imagenet-21k pretraining for the masses.
\newblock In \emph{Thirty-fifth Conference on Neural Information Processing
  Systems Datasets and Benchmarks Track (Round 1)}, 2021.

\bibitem[Touvron et~al.(2023)Touvron, Lavril, Izacard, Martinet, Lachaux,
  Lacroix, Rozi{\`e}re, Goyal, Hambro, Azhar, et~al.]{touvron2023llama}
Hugo Touvron, Thibaut Lavril, Gautier Izacard, Xavier Martinet, Marie-Anne
  Lachaux, Timoth{\'e}e Lacroix, Baptiste Rozi{\`e}re, Naman Goyal, Eric
  Hambro, Faisal Azhar, et~al.
\newblock Llama: Open and efficient foundation language models.
\newblock \emph{arXiv preprint arXiv:2302.13971}, 2023.

\bibitem[Van~Horn et~al.(2015)Van~Horn, Branson, Farrell, Haber, Barry,
  Ipeirotis, Perona, and Belongie]{van2015building}
Grant Van~Horn, Steve Branson, Ryan Farrell, Scott Haber, Jessie Barry, Panos
  Ipeirotis, Pietro Perona, and Serge Belongie.
\newblock Building a bird recognition app and large scale dataset with citizen
  scientists: The fine print in fine-grained dataset collection.
\newblock In \emph{Proceedings of the IEEE Conference on Computer Vision and
  Pattern Recognition (CVPR)}, 2015.

\bibitem[Wah et~al.(2011)Wah, Branson, Welinder, Perona, and
  Belongie]{wah2011caltech}
Catherine Wah, Steve Branson, Peter Welinder, Pietro Perona, and Serge
  Belongie.
\newblock The caltech-ucsd birds-200-2011 dataset.
\newblock \emph{California Institute of Technology}, 2011.

\bibitem[Wang et~al.(2024)Wang, Dedhia, and Jha]{wang2024zero}
Hongjie Wang, Bhishma Dedhia, and Niraj~K Jha.
\newblock Zero-tprune: Zero-shot token pruning through leveraging of the
  attention graph in pre-trained transformers.
\newblock In \emph{Proceedings of the IEEE/CVF Conference on Computer Vision
  and Pattern Recognition}, pages 16070--16079, 2024.

\bibitem[Xin et~al.(2024{\natexlab{a}})Xin, Du, Wang, Lin, and Yan]{xin2024vmt}
Yi Xin, Junlong Du, Qiang Wang, Zhiwen Lin, and Ke Yan.
\newblock Vmt-adapter: Parameter-efficient transfer learning for multi-task
  dense scene understanding.
\newblock In \emph{Proceedings of the AAAI Conference on Artificial
  Intelligence (AAAI)}, 2024{\natexlab{a}}.

\bibitem[Xin et~al.(2024{\natexlab{b}})Xin, Du, Wang, Yan, and
  Ding]{xin2023mmap}
Yi Xin, Junlong Du, Qiang Wang, Ke Yan, and Shouhong Ding.
\newblock Mmap : Multi-modal alignment prompt for cross-domain multi-task
  learning.
\newblock In \emph{Proceedings of the AAAI Conference on Artificial
  Intelligence (AAAI)}, 2024{\natexlab{b}}.

\bibitem[Xin et~al.(2024{\natexlab{c}})Xin, Luo, Liu, Zhou, Cheng, Lee, Du,
  Wang, Chen, Liu, et~al.]{xin2024v}
Yi Xin, Siqi Luo, Xuyang Liu, Haodi Zhou, Xinyu Cheng, Christina~E Lee, Junlong
  Du, Haozhe Wang, MingCai Chen, Ting Liu, et~al.
\newblock V-petl bench: A unified visual parameter-efficient transfer learning
  benchmark.
\newblock \emph{Advances in neural information processing systems},
  37:\penalty0 80522--80535, 2024{\natexlab{c}}.

\bibitem[Xin et~al.(2024{\natexlab{d}})Xin, Luo, Zhou, Du, Liu, Fan, Li, and
  Du]{xin2024parameter}
Yi Xin, Siqi Luo, Haodi Zhou, Junlong Du, Xiaohong Liu, Yue Fan, Qing Li, and
  Yuntao Du.
\newblock Parameter-efficient fine-tuning for pre-trained vision models: A
  survey.
\newblock \emph{arXiv preprint arXiv:2402.02242}, 2024{\natexlab{d}}.

\bibitem[Xin et~al.(2025)Xin, Yan, Qin, Li, Liu, Li, Huang, Zhou, Zhang, Zhuo,
  et~al.]{xin2025lumina}
Yi Xin, Juncheng Yan, Qi Qin, Zhen Li, Dongyang Liu, Shicheng Li, Victor
  Shea-Jay Huang, Yupeng Zhou, Renrui Zhang, Le Zhuo, et~al.
\newblock Lumina-mgpt 2.0: Stand-alone autoregressive image modeling.
\newblock \emph{arXiv preprint arXiv:2507.17801}, 2025.

\bibitem[Zaken et~al.(2022)Zaken, Ravfogel, and Goldberg]{zaken2021bitfit}
Elad~Ben Zaken, Shauli Ravfogel, and Yoav Goldberg.
\newblock Bitfit: Simple parameter-efficient fine-tuning for transformer-based
  masked language-models.
\newblock \emph{Proceedings of the Annual Meeting of the Association for
  Computational Linguistics (ACL)}, 2022.

\bibitem[Zhai et~al.(2019)Zhai, Puigcerver, Kolesnikov, Ruyssen, Riquelme,
  Lucic, Djolonga, Pinto, Neumann, Dosovitskiy, et~al.]{zhai2019large}
Xiaohua Zhai, Joan Puigcerver, Alexander Kolesnikov, Pierre Ruyssen, Carlos
  Riquelme, Mario Lucic, Josip Djolonga, Andre~Susano Pinto, Maxim Neumann,
  Alexey Dosovitskiy, et~al.
\newblock A large-scale study of representation learning with the visual task
  adaptation benchmark.
\newblock \emph{arXiv preprint arXiv:1910.04867}, 2019.

\bibitem[Zhang et~al.(2024{\natexlab{a}})Zhang, Zhou, and Liu]{zhang2022neural}
Yuanhan Zhang, Kaiyang Zhou, and Ziwei Liu.
\newblock Neural prompt search.
\newblock \emph{IEEE Transactions on Pattern Analysis and Machine Intelligence
  (TPAMI)}, 2024{\natexlab{a}}.

\bibitem[Zhang et~al.(2024{\natexlab{b}})Zhang, Zhang, Gao, Zhang, Shutova,
  Zhou, and Zhang]{zhang2024gradient}
Zhi Zhang, Qizhe Zhang, Zijun Gao, Renrui Zhang, Ekaterina Shutova, Shiji Zhou,
  and Shanghang Zhang.
\newblock Gradient-based parameter selection for efficient fine-tuning.
\newblock In \emph{Proceedings of the IEEE Conference on Computer Vision and
  Pattern Recognition (CVPR)}, 2024{\natexlab{b}}.

\end{thebibliography}
